\documentclass[runningheads]{llncs}
\usepackage{graphicx,wrapfig}
\usepackage{amsmath,amssymb} 
\usepackage{color}
\usepackage[hidelinks]{hyperref}

\begin{document}
\title{SDC-Net: Video prediction using spatially-displaced convolution} 

\titlerunning{SDC-Net: Video prediction using spatially-displaced convolution}
%
\author{Fitsum A. Reda\and Guilin Liu\and Kevin J. Shih\and Robert Kirby\and Jon Barker\and\\ David Tarjan\and Andrew Tao\and Bryan Catanzaro}
%
\authorrunning{F.A. Reda et al.}
%
\institute{Nvidia Corporation, Santa Clara CA 95051, USA \\
}
\maketitle              
\begin{figure}
    \centering
    \includegraphics[width=\textwidth]{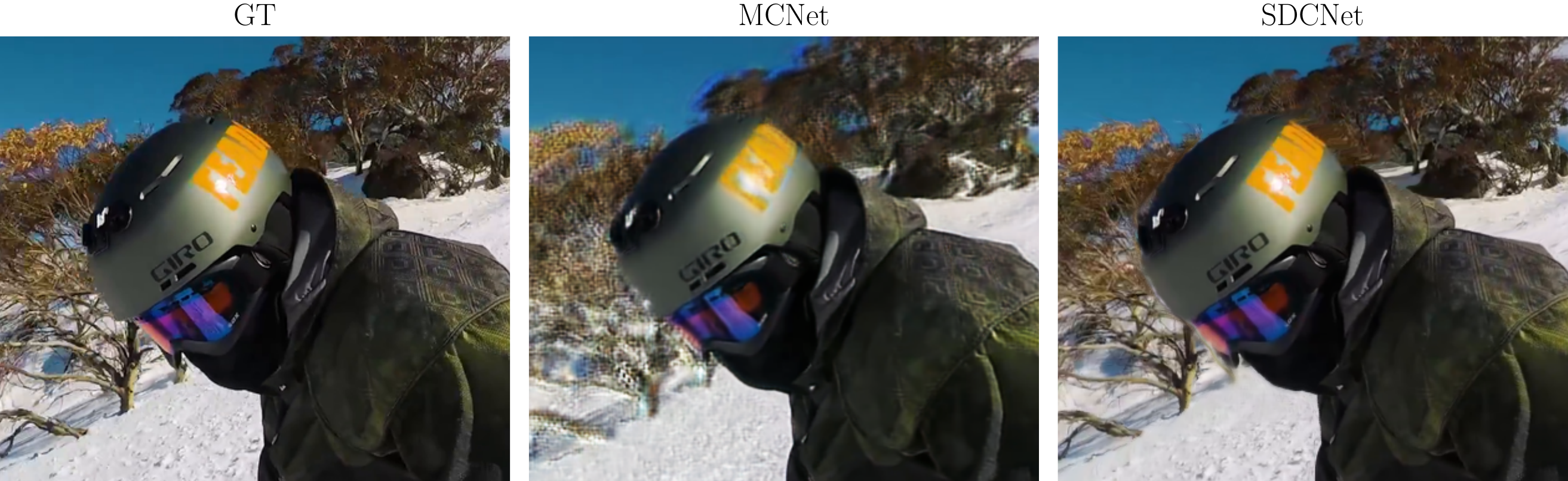}
    \caption{Frame prediction on a YouTube video frame featuring a panning camera. Left to right: Ground-truth, MCNet~\cite{decomposing_motion_content} result, and our SDC-Net result. The SDC-Net predicted frame is sharper and preserves fine image details, while color distortion and blurriness is seen in the tree and text in MCNet's predicted frame. Project \textcolor{magenta}{\href{https://nv-adlr.github.io/publication/2018-SDCNet}{website}}.}
    \label{fig:teaser}
\end{figure}
\begin{abstract}
We present an approach for high-resolution video frame prediction by conditioning on both past frames and past optical flows. Previous approaches rely on resampling past frames, guided by a learned future optical flow, or on direct generation of pixels. Resampling based on flow is insufficient because it cannot deal with disocclusions. Generative models currently lead to blurry results. Recent approaches synthesis a pixel by convolving input patches with a predicted kernel. However, their memory requirement increases with kernel size. Here, we present \emph{spatially-displaced convolution} (SDC) module for video frame prediction. We learn a motion vector and a kernel for each pixel and synthesize a pixel by applying the kernel at a displaced location in the source image, defined by the predicted motion vector. Our approach inherits the merits of both vector-based and kernel-based approaches, while ameliorating their respective disadvantages. We train our model on 428K unlabelled 1080p video game frames. Our approach produces state-of-the-art results, achieving an SSIM score of 0.904 on high-definition YouTube-8M videos, 0.918 on Caltech Pedestrian videos. Our model handles large motion effectively and synthesizes crisp frames with consistent motion. Codes available at \textcolor{magenta}{ \href{https://github.com/NVIDIA/semantic-segmentation/tree/sdcnet/sdcnet}{NVIDIA/semantic-segmentation/sdcnet}}.
\end{abstract}
\section{Introduction}
Video prediction is the task of inferring future frames from a sequence of past frames. The ability to predict future frames could find applications in various domains -- ranging from future state estimation for self-driving vehicles to video analysis. For a video prediction model to perform well, it must accurately capture not only how objects move, but also how their displacement affects the visibility and appearance of surrounding structures. Our work focuses on predicting one or more immediate next frames that are sharp, realistic and at high resolution. 

Another attribute of the video prediction task is that models can be trained on raw unlabeled video frames. We train our models on large amounts of high resolution footage from video game sequences, which we find improves accuracy because video game sequences contain a large range of motion. We demonstrate that the resulting models perform well not only on video game footage, but also on real-life footage.

Video prediction is an active research area and our work builds on the literature ~\cite{luc2017predicting,vukotic2017one,van2017transformation,leibfried2016deep,mahjourian2017geometry,vondrick2016generating,denton2018stochastic,babaeizadeh2017stochastic,byeon2017fully,oliu2017folded,lu2017flexible}.
Previous approaches for video prediction often focus on direct synthesis of pixels using generative models. For instance, convolutional neural networks were used to predict pixel RGB values, while recurrent mechanisms were used to model temporal changes. Ranzato et al.~\cite{ranzato2014video} proposed to partition input sequences into a dictionary of image patch centroids and trained recurrent neural networks (RNN) to generate target images by indexing the dictionaries. Srivastava et al. ~\cite{srivastava15_unsup_video} and Villegas et al.~\cite{decomposing_motion_content} used a convolutional Long-Short-Term-Memory (LSTM) encoder-decoder architecture conditioned on previous frame data. Similarly, Lotter et al.~\cite{predictive_coding} presented a predictive coding RNN architecture to model the motion dynamics of objects in the image for frame prediction. Mathieu et al.~\cite{Mathieu2016BeyondMSE} proposed a multi-scale conditional generative adversarial network (GAN) architecture to alleviate the short range dependency of single-scale architectures. These approaches, however, suffer from blurriness and do not model large object motions well. This is likely due to the difficulty in directly regressing to pixel values, as well as the low resolution and lack of large motion in their training data.  

Another popular approach for frame synthesis is learning to transform input frames. Liang et al.~\cite{Liang2017DualGan} proposed a generative adversarial network (GAN) approach with a joint future optical-flow and future frame discriminator. However, ground truth optical flows are not trivial to collect at large scale. Training with estimated optical flows could also lead to erroneous supervision signals. Jiang et al.~\cite{jiang2017super} presented a model to learn offset vectors for sampling for frame interpolation, and perform frame synthesis using bilinear interpolation guided by the learned sampling vectors. These approaches are desirable in modeling large motion. However, in our experiments, we found sampling vector-based synthesis results are often affected by speckled noise. 
 
One particular approach proposed by Niklaus et al.~\cite{separable_convolution,adaptive_convolution} and Vondrick et al.~\cite{diff_transformer} for frame synthesis is to learn to predict sampling kernels that adapt to each output pixel. A pixel is then synthesized as the weighted sampling of a source patch centered at the pixel location. Niklaus et al.~\cite{separable_convolution,adaptive_convolution} employed this for the related task of video frame interpolation, applying predicted sampling kernels to consecutive frames to synthesize the intermediate frame. In our experiments, we found the kernel-based approaches to be effective in keeping objects intact as they are transformed. However, this approach cannot model large motion, since its displacement is limited by the kernel size. Increasing kernel size can be prohibitively expensive.

Inspired by these approaches, we present a spatially-displaced convolutional (SDC) module for video frame prediction. We learn a motion vector and a kernel for each pixel and synthesize a pixel by applying the kernel at a displaced location in a source image, defined by the predicted motion vector. Our approach inherits the merits of both sampling vector-based and kernel-based approaches, while ameliorating their respective disadvantages. We take the large-motion advantage of sampling vector-based approach, while reducing the speckled noise patterns. We take the clean object boundary prediction advantages of the kernel-based approaches, while significantly reducing kernel sizes, hence reducing the memory demand. 

The contributions of our work are:
\begin{itemize}
    \item We propose a deep model for high-resolution frame prediction from a sequence of past frames.
    \item We propose a spatially-displaced convolutional (SDC) module for effective frame synthesis via transformation learning.
    \item We compare our SDC module with kernel-based, vector-based and  state-of-the-art approaches.
\end{itemize}
\section{Methods}
Given a sequence of frames $\textbf{I}_{1:t}$ (the immediate past $t$ frames), our work aims to predict the next future frame ${\textbf{I}_{t+1}}$. We formulate the problem as a transformation learning problem
\begin{equation} \label{eq:1}
{\textbf{I}_{t+1}} = \mathcal{T}\Big(\mathcal{G}\big(\textbf{I}_{1:t}\big), \textbf{I}_{1:t}\Big) ,
\end{equation}
where $\mathcal{G}$ is a learned function that predicts transformation parameters, and $\mathcal{T}$ is a transformation function. In prior work, $\mathcal{T}$ can be a bilinear sampling operation guided by a motion vector~\cite{jiang2017super,liu2017video}:
\begin{equation} \label{eq:bilinear_model}
\textbf{I}_{t+1}(x,y)= f\big(\textbf{I}_{t}(x+u,y+v)\big) , 
\end{equation}
where $f$ is a bilinear interpolator~\cite{liu2017video}, $(u,v)$ is a motion vector predicted by $\mathcal{G}$, and $\textbf{I}_{t}(x,y)$ is a pixel value at $(x,y)$ in the immediate past frame $\textbf{I}_{t}$. We refer this approach as a vector-based resampling. Fig. \ref{fig:sampling_approaches}a illustrates this approach.

An alternative approach is to define $\mathcal{T}$ as a convolution module that combines motion or displacement learning and resampling into a single operation~\cite{separable_convolution,adaptive_convolution,diff_transformer}:
\begin{equation} \label{eq:conv_model}
\textbf{I}_{t+1}(x,y)=\mathrm{K}(x,y)*\textbf{P}_{t}(x,y), 
\end{equation}
where $\mathrm{K}(x,y) \in \mathrm{R}^{\mathrm{N}\times \mathrm{N}}$ is an N$\times$N 2D kernel predicted by $\mathcal{G}$ at $(x,y)$ and $\textbf{P}_t(x,y)$ is an N$\times$N patch centered at $(x,y)$ in $\textbf{I}_t$. We refer this approach as adaptive kernel-based resampling~\cite{separable_convolution,adaptive_convolution}. Fig. \ref{fig:sampling_approaches}b illustrates this approach.

Since equation (\ref{eq:bilinear_model}) considers few pixels in synthesis, its results often appear degraded by speckled noise patterns. It can, however, model large displacements without a significant increase in parameter count. On the other hand, equation (\ref{eq:conv_model}) produces visually pleasing results for small displacements, but requires large kernels to be predicted at each location to capture large motions. As such, the kernel-based approach can easily become not only costly at inference, but also difficult to train.

\subsection{Spatially Displaced Convolution}
To achieve the best of both worlds, we propose a hybrid solution -- the \emph{Spatially Displaced Convolution} (SDC). The SDC uses predictions of both a motion vector $(u,v)$ and an adaptive kernel $\mathrm{K}(x,y)$, but convolves the predicted kernel with a patch at the displaced location $(x+u, y+v)$ in $\textbf{I}_t$.
Pixel synthesis using SDC is computed as:
\begin{equation} \label{eq:displaced_convolution}
\textbf{I}_{t+1}(x,y)=\mathrm{K}(x,y)*\textbf{P}_{t}(x+u,y+v).
\end{equation}
The predicted pixel $\textbf{I}_{t+1}(x,y)$ is thus the weighted sampling of pixels in an N$\times$N region centered at $(x+u, y+v)$ in $\textbf{I}_{t}$. The patch $\textbf{P}_t(x+u,y+v)$ is bilinearly sampled at non-integral coordinates. Fig. \ref{fig:sampling_approaches}c illustrates our SDC-based approach.

Setting $K(x,y)$ to a kernel of all-zeros except for a one at the center reduce the SDC to equation (\ref{eq:bilinear_model}), whereas setting $u$ and $v$ to zero reduces it to equation (\ref{eq:conv_model}). However, it is important to note that the SDC is not the same as applying equation (\ref{eq:bilinear_model}) and equation (\ref{eq:conv_model}) in succession. If applied in succession, the N$\times$N patch sampled by $K(x,y)$ would be subject to the resampling effect of equation (\ref{eq:bilinear_model}) as opposed to being the original patch from $\textbf{I}_t$. 
\begin{figure}[t!]
    \centering
    \scalebox{1.0}{\includegraphics[width=1.0\textwidth,trim=0 135 0 150,clip]{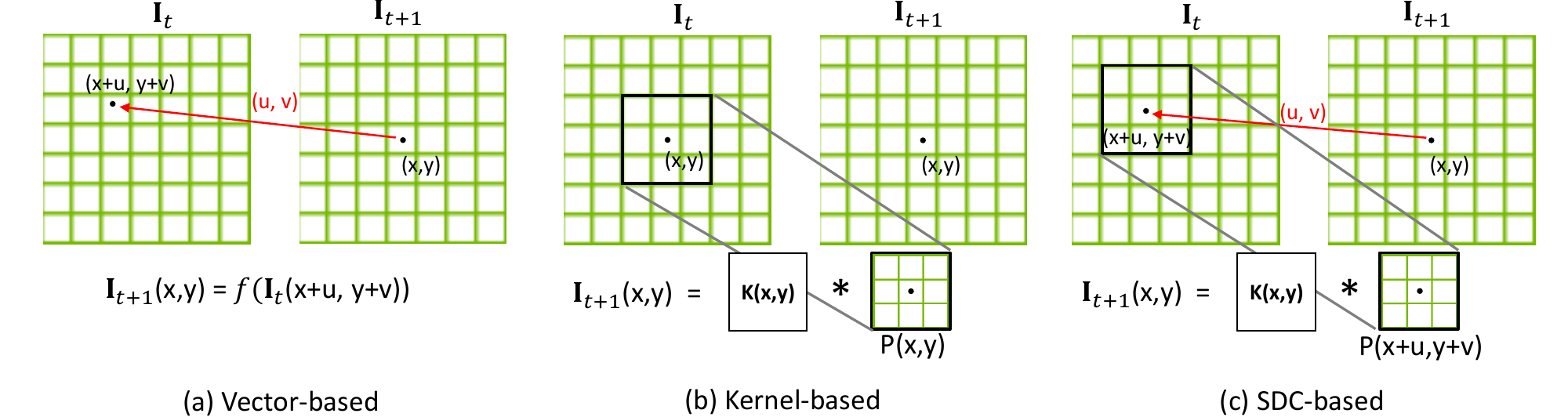}}
    \caption{Illustration of sampling-based pixel synthesis. (a) Vector-based with a bilinear interpolation. (b) Kernel-based, a convolution with a centered patch. (c) our SDC-based method, a convolution with a displaced patch.}
    \label{fig:sampling_approaches}
\end{figure}

Our SDC effectively decouples displacement and kernel learning, allowing us to achieve the visually pleasing results of kernel-based approaches while keeping the kernel sizes small. We also adopt separable kernels~\cite{separable_convolution} for $\mathrm{K}(x,y)$ to further reduce computational cost. At each location, we predict a pair of 1D kernels and calculate the outer-product of them to form a 2D kernel. This reduces our kernel parameter count from $\mathrm{N}^2$ to $2\mathrm{N}$. In total, our model predicts $2\mathrm{N}+2$ parameters for each pixel, including the motion vector. We empirically set $N=11$. Inference at 1080p resolution uses 174MB of VRAM, which easily fits in GPU memory. 

We develop deep neural networks to learn motion vectors and kernels adapted to each output pixel. The SDC is fully differentiable and thus allows our model to train end-to-end. Losses for training are applied to the SDC-predicted frame. We also condition our model on both past frames and past optical flows. This allows our network to easily capture motion dynamics and evolution of pixels needed to learn the transformation parameters. We formulate our model as: 

\begin{equation} \label{eq:5}
{\textbf{I}_{t+1}} = \mathcal{T}\Big(\mathcal{G}\big(\textbf{I}_{1:t},\textbf{F}_{2:t}\big), \textbf{I}_{t}\Big) ,
\end{equation}
where transformation $\mathcal{T}$ is realized with SDC and operates on the most recent input $\textbf{I}_{t}$, and $\textbf{F}_{i}$ is the backwards optical flow (see Section \ref{sec:optical_flow} ) between $\textbf{I}_{i}$ and $\textbf{I}_{i-1}$. We calculate \textbf{F} using state-of-the-art neural network-based optical flow models~\cite{dosovitskiy2015flownet,ilg2017flownet,sun2017pwc}.

Our approach naturally extends to multiple frame prediction ${\textbf{I}_{t+1:t+D}}$ by recursively re-circulating SDC predicted frames back as inputs. For instance, to predict a frame two steps ahead, we re-circulate the SDC predicted frame ${\textbf{I}_{t+1}}$ as input to our model to predict ${\textbf{I}_{t+2}}$.  

\subsection{Network Architecture}
We realize $\mathcal{G}$ using a fully convolutional network. Our model takes in a sequence of past frames $\textbf{I}_{1:t}$ and past optical flows $\textbf{F}_{2:t}$ and outputs pixel-wise separable kernels $\{\mathrm{K}_{u}$, $\mathrm{K}_{v}\}$ and a motion vector $(u,v)$. We use 3D convolutions to convolve across width, height, and time. We concatenate RGB channels from our input images to the two optical flow channels to create 5 channels per frame. The topology of our architecture gets inspiration from various V-net type typologies ~\cite{dosovitskiy2015flownet,milletari2016v,ronneberger2015u}, with an encoder and a decoder. Each layer of the encoder applies 3D convolutions followed by a Leaky Rectified Unit (LeakyRELU)~\cite{he2015delving} and a convolution with a stride (1,2,2) to downsample features to capture long-range spatial dependencies. Following \cite{dosovitskiy2015flownet}, we use 3x3x3 convolution kernels, except for the first and second layers where we use 3x7x7 and 3x5x5 for capturing large displacements. Each decoder sub-part applies deconvolutions~\cite{long2015fully} followed by LeakyRELU, and a convolution after corresponding features from the contracting part have been concatenated. The decoding part also has several heads, one head for $(u,v)$ and one each for $\mathrm{K}_{u}$ and $\mathrm{K}_{v}$. The last two decoding layers of $\mathrm{K}_{u}$ and $\mathrm{K}_{v}$ use upsampling with a trilinear mode, instead of normal deconvolution, to minimize the checkerboard effect~\cite{odena2016deconvolution}. Finally, we apply repeated convolutions in each decoding head to reduce the time dimension to 1.
\begin{figure}
    \centering
    \scalebox{0.9}{\includegraphics[width=1.0\textwidth]{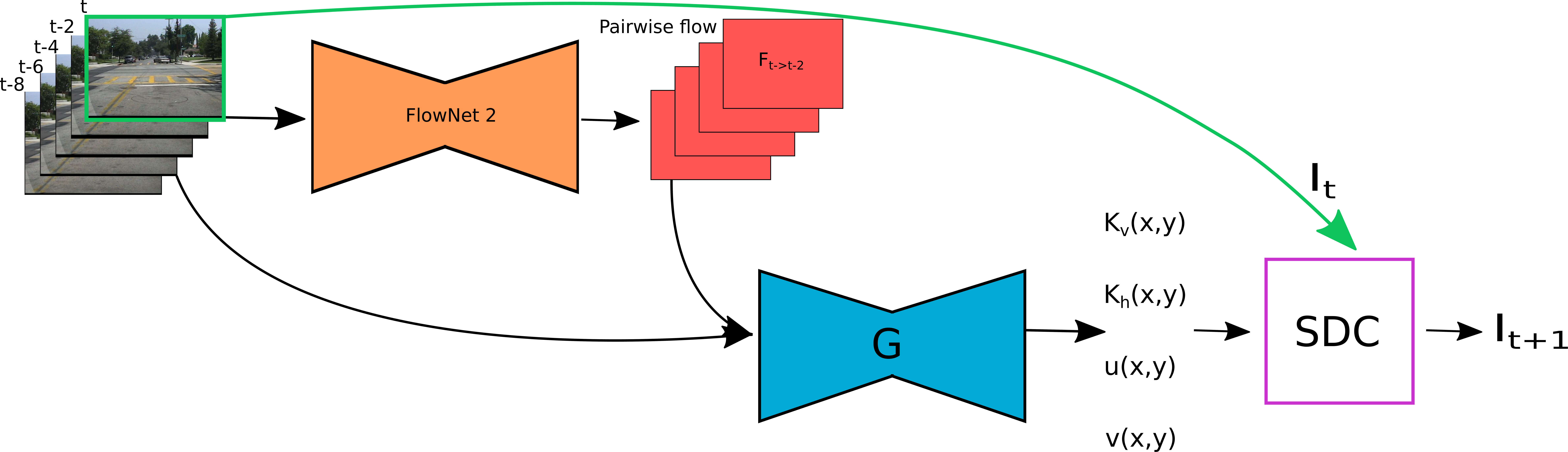}}
    \caption{Our model $\mathcal{G}$ takes in a frame sequence and pairwise flow estimates as input, and returns parameters for the SDC module to  transform $\textbf{I}_t$ to $\textbf{I}_{t+1}$.}
    \label{fig:sampling_approaches}
\end{figure}
\subsection{Optical Flow}
\label{sec:optical_flow}

We calculate the inter-frame optical flow we input to our model $\mathcal{G}$ using FlowNet2~\cite{ilg2017flownet}, a state-of-the-art optical flow model. This allows us to extrapolate motion conditioned on past flow information.
We calculate backwards optical flows because we model our transformation learning problem with backwards resampling, i.e. predict a sampling location in $\textbf{I}_t$ for each location in $\textbf{I}_{t+1}$.

\begin{figure}
    \centering
    \scalebox{0.6}{\includegraphics[width=0.9\textwidth,trim=0 280 354 0,clip]{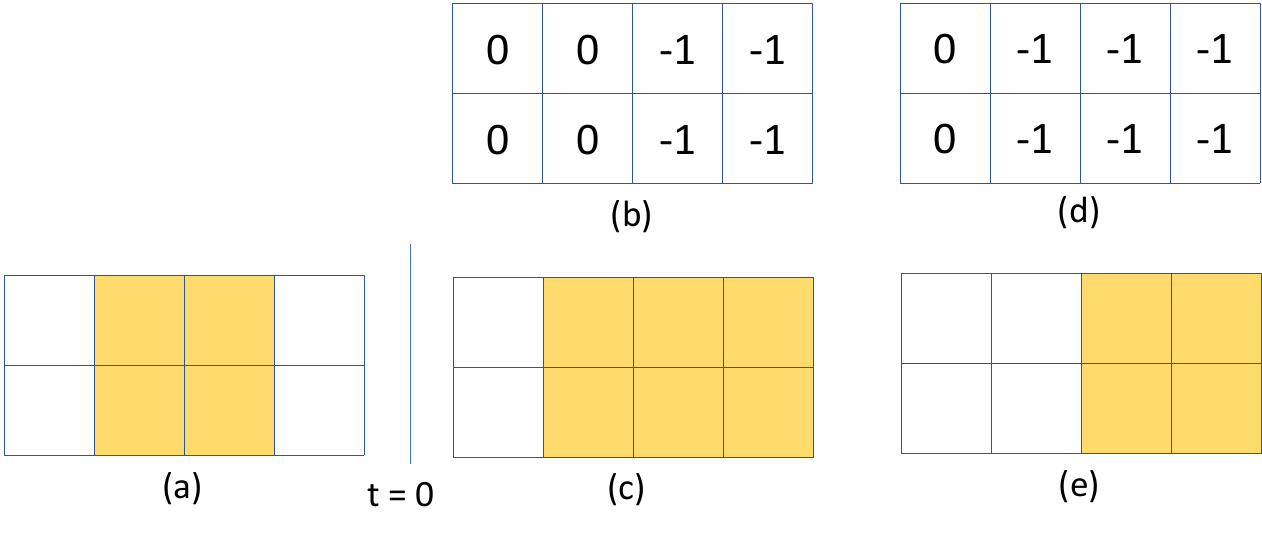}}
    \caption{Disocclusion illustration using backwards optical-flow. Values in top-row indicate vector magnitude in the horizontal axis. (a) frame at $t=0$; (b) optical flow at $t=1$; (c) output of resampling (a) using (b); (d) correct sampling vectors; and (e) resampling of (a) using (d). A direct use of optical-flow for frame prediction leads to undesirable foreground stretching in dis-occluded pixels.}
    \label{fig:optical_Flow}
\end{figure}

It is important to note that the motion vectors $(u,v)$ predicted by our model $\mathcal{G}$ at each pixel are not equivalent to optical flow vectors $\textbf{F}_{t+1}$, as pure backwards optical flow is undefined (or zero valued) for dis-occluded pixels (pixels not visible in the previous frame due to occlusion). A schematic explanation of the disocclusion problem is shown in Fig. \ref{fig:optical_Flow}, where a 2$\times$2 square is moving horizontally to the right at a speed of 1 pixel per step. The ground-truth backward optical flow at $t=1$ is shown in Fig. \ref{fig:optical_Flow}b. As shown in Fig. \ref{fig:optical_Flow}c, resampling the square at $t=0$ using the perfect optical flow will duplicate the left border of the square because the optical flow is zero at the second column. To achieve a perfect synthesis via resampling at $t=1$, as shown in Fig. \ref{fig:optical_Flow}e, adaptive sampling vectors must be used. Fig. \ref{fig:optical_Flow}d shows an example of such sampling vectors, in which a $-1$ is used to fill-in dis-occluded region. A learned approach is necessary here as it not only allows the disocclusion sampling to adapt for various degrees of motion, but also allows for a learned solution for which background pixels from the previous frame would look best in the filled gap.
\subsection{Loss Functions}
Our primary loss function is the L1 loss over the predicted image: ${\mathcal{L}}_{1}={\left\lVert\textbf{I}_{t+1}-\textbf{I}_{t+1}^{g}\right\rVert}_{1}$, where $\textbf{I}_{i}^g$ is a target and $\textbf{I}_{i}$ is a predicted frame. We found the L1 loss to be better at capturing small changes compared to L2, and generally produces sharper images.

We also incorporated the L1 norm on high-level VGG-16 feature representations~\cite{Simonyan14c}. Specifically, we used the perceptual and style loss~\cite{johnson2016perceptual}, defined as:
\begin{equation} \label{eq:perceptual}
{\mathcal{L}_{perceptual}} = \sum_{l=1}^{\mathrm{L}}{\kappa_l {\left\lVert{\Psi}_{l}(\textbf{I}_{t+1})-{\Psi}_{l}(\textbf{I}_{t+1}^{g})\right\rVert}_{1} }, 
\end{equation}
and
\begin{equation} \label{eq:style}
{\mathcal{L}_{style}} = \sum_{l=1}^{\mathrm{L}}{ \kappa_l{\left\lVert{{\big(\Psi}_{l}(\textbf{I}_{t+1})\big)}^{\intercal}{{\big(\Psi}_{l}(\textbf{I}_{t+1})\big)} - {\big({\Psi}_{l}(\textbf{I}_{t+1}^{g})\big)}^{\intercal}{\big({\Psi}_{l}(\textbf{I}_{t+1}^{g})\big)}\right\rVert}_{1} }. 
\end{equation}
Here, ${\Psi}_{l}(\textbf{I}_{i})$ is the feature map from the ${l}$th selected layer of a pre-trained Imagenet VGG-16 for $\textbf{I}_{i}$, $\mathrm{L}$ is the number of layers considered, and $\kappa_l$ is a normalization factor $1/C_lH_lK_l$ (channel, height, width) for the $l$th selected layer. We use the perceptual and style loss terms in conjunction with the L1 over RGB as follows:
\begin{equation}\label{eq:finetune}
    \mathcal{L}_{finetune}=w_l\mathcal{L}_1+w_s\mathcal{L}_{style}+w_p\mathcal{L}_{perceptual}.
\end{equation}
We found the finetune loss to be robust in eliminating the checkerboard artifacts and generates a much sharper prediction than L1 alone.

Finally, we introduce a loss to initialize the adaptive kernels, which we found to significantly speed up training. We use the L2 norm to initialize kernels $\mathrm{K}_{u}$ and $\mathrm{K}_{v}$ as a middle-one-hot vector each. That is, all elements in each kernel are set very close to zero, except for the middle element which is initialized close to one. When $\mathrm{K}_{u}$ and $\mathrm{K}_{v}$ elements are initialized as middle-hot vectors, the output of our displaced convolution described in equation (\ref{eq:displaced_convolution}) will be the same as our vector-based approach described in equation (\ref{eq:bilinear_model}). The kernel loss is expressed as:

\begin{equation} \label{eq:kernel_loss}
{\mathcal{L}_{kernel}} = \sum_{x=1}^{\mathrm{W}}{\sum_{y=1}^{\mathrm{H}}{\Big(\left\lVert\mathrm{K}_{u}(x,y) - \textbf{1}^{<\frac{\mathrm{N}}{2}>}\right\rVert_{2}^{2}+
\left\lVert \mathrm{K}_{v}(x,y) -\textbf{1}^{<\frac{\mathrm{N}}{2}>} \right\rVert_{2}^{2}
\Big)}}, 
\end{equation}
where $\textbf{1}^{<\frac{\mathrm{N}}{2}>}$ is a middle-one-hot vector, and $\mathrm{W}$ and $\mathrm{H}$ are the width and height of images.

Other loss functions considered include the L1 or L2 distance between predicted motion vectors $(u,v)$ and target optical flows. We found this loss to lead to inferior results. As discussed in Section \ref{sec:optical_flow}, optimizing for optical flow will not properly handle the disocclusion problem. Further, use of estimated optical flow as a training target introduces additional noise.

\subsection{Training}
We trained our SDC model using frames extracted from many short sequence videos. To allow our model to learn robust invariances, we selected frames in high-definition video game plays with realistic, diverse content, and a wide range of motion. We collected 428K 1080p frames from GTA-V and Battlefield-1 game plays. Each example consists of five ($t$=5) consecutive 256$\times$256 frames randomly cropped from the full-HD sequence. We use a batch size of 128 over 8 V100 GPUs. 

We optimize with Adam~\cite{kingma2014adam} using ${\beta}_{1}=0.9$, and ${\beta}_{2}=0.999$ with no weight decay. 
First, we optimize our model to learn $(u,v)$ using ${\mathcal{L}}_{1}$ loss with a learning rate of ${1e}^{-4}$ for 400 epochs. Optimizing for $(u,v)$ alone allows our network to capture large and coarse motions faster. Next, we fix all weights of the network except for the decoding heads of ${K}_{u}$ and ${K}_{v}$ and train them using our $\mathcal{L}_{kernel}$ loss defined in equation (\ref{eq:kernel_loss}) to initialize kernels at each output pixel as middle-one-hot vectors. Then, we optimize all weights in our deep model using ${\mathcal{L}}_{1}$ loss and a learning rate of ${1e}^{-5}$ for 300 epochs to jointly fine-tune the $(u,v)$ and (${K}_{u}$, ${K}_{v}$) at each pixel. Since we optimize for both kernels and motion vectors in this step, our network learns to pick up small and subtle motions and corrects disocclusion related artifacts. Finally, we further fine-tune all weights in our model using $\mathcal{L}_{finetune}$ at a learning rate of ${1e}^{-5}$. The weights we use to combine losses are 0.2, 0.06, 36.0 for $w_l$, $w_p$, and $w_s$ respectively. We used the activations from VGG-16 layers \texttt{relu1\_2}, \texttt{relu2\_2} and \texttt{relu3\_3} for the perceptual and style loss terms. The last fine-tuning step of our training makes predictions sharper and produces visually appealing frames in our video prediction task. We initialized the FlowNet2 model with pre-trained weights\footnote{\texttt{https://github.com/lmb-freiburg/flownet2}} and fix them during training. 

\section{Experiments}

We implemented all our Vector, Kernel, and SDC-based models using PyTorch~\cite{paszke2017automatic}. To efficiently train our model, we wrote a CUDA custom layer for our SDC module. We set kernel size to 51$\times$51 for the Kernel-based model as suggested in~\cite{separable_convolution}. The kernel size for our SDC-based model is 11$\times$11. Inference using our SDC-based model at 1080p takes 1.66sec, of which 1.03 sec is spent on FlowNet2.
\subsection{Datasets and Metrics}
We considered two classes of video datasets that contain complex real world scenes: Caltech Pedestrian~\cite{Dollar2012PAMI,dollarCVPR09peds} (CaltechPed) car-mounted camera videos and 26 high definition videos collected from YouTube-8M~\cite{abu2016youtube}. 

We used metrics L1, Mean-Squared-Error (MSE/L2)~\cite{predictive_coding}, Peak-Signal-To-Noise (PSNR), and Structural-Similarity-Image-Metric (SSIM)~\cite{wang2004image} to evaluate quality of prediction. Higher values of SSIM and PSNR indicate better quality.

\subsection{Comparison on low-quality videos}
\begin{wraptable}{r}{5cm}
\caption{Next frame prediction accuracy on Caltech Pedestrian~\cite{Dollar2012PAMI,dollarCVPR09peds}.  L2 results are in 1e-3.}\label{table:1}.
\centering
\begin{tabular}{ p{2.8cm} p{0.9cm} p{1cm} } 
\hline
Methods & L2 & SSIM \\
\hline 
 BeyondMSE\cite{Mathieu2016BeyondMSE} & 3.42 &   0.847  \\
 PredNet\cite{predictive_coding} & 3.13 &   0.884  \\
 MCNet\cite{decomposing_motion_content} & 2.50 & 0.879  \\
 DualGAN\cite{Liang2017DualGan} & 2.41 &   0.899  \\ 
 CopyLast & 5.84 &   0.811  \\ 
 Our Vector-based & 2.47 &   0.902  \\ 
 Our Kernel-based & 2.19 &   0.896  \\ 
 Our SDC-based & \textbf{1.62} &   \textbf{0.918}  \\
\hline
\end{tabular}
\end{wraptable} 
Table \ref{table:1} presents next frame prediction comparisions with BeyondMSE~\cite{Mathieu2016BeyondMSE}, PredNet~\cite{predictive_coding}, MCNet~\cite{decomposing_motion_content}, and DualGAN~\cite{Liang2017DualGan} on CaltechPed test partition. We also compare with CopyLast, which is the trivial baseline that uses the most recent past frame as the prediction. For PredNet and DualGAN, we directly report results presented in~\cite{predictive_coding} and~\cite{liu2017video}, respectively.

For BeyondMSE\footnote{\texttt{ https://github.com/coupriec/VideoPredictionICLR2016}} and MCNet\footnote{\texttt{ https://github.com/rubenvillegas/iclr2017mcnet}}, we evaluated using released pre-trained models. 

Our SDC-based model outperforms all other models, achieving an L2 score of ${1.62}\times{10}^{-3}$ and SSIM of $0.918$, compared to the state-of-the-art DualGAN model which has an L2 score of ${2.41\times{10}^{-3}}$ and SSIM of ${0.899}$. The MCNet which was trained on dataset that is equally as large as ours shows inferior results with L2 of ${2.50\times{10}^{-3}}$ and SSIM of $0.879$. CopyLast method has significantly worse L2 of $5.84\times{10}^{-3}$ and SSIM of ${0.811}$, making it a significantly less viable approach for next frame prediction. Our Vector-based approach has higher accuracy than our Kernel-based approach. Since the CaltechPed videos contain slightly larger motion, the Vector-based approach, which is advantageous in large motion sequences, is expected to perform better.

\begin{figure}
    \centering
    \includegraphics[width=\textwidth]{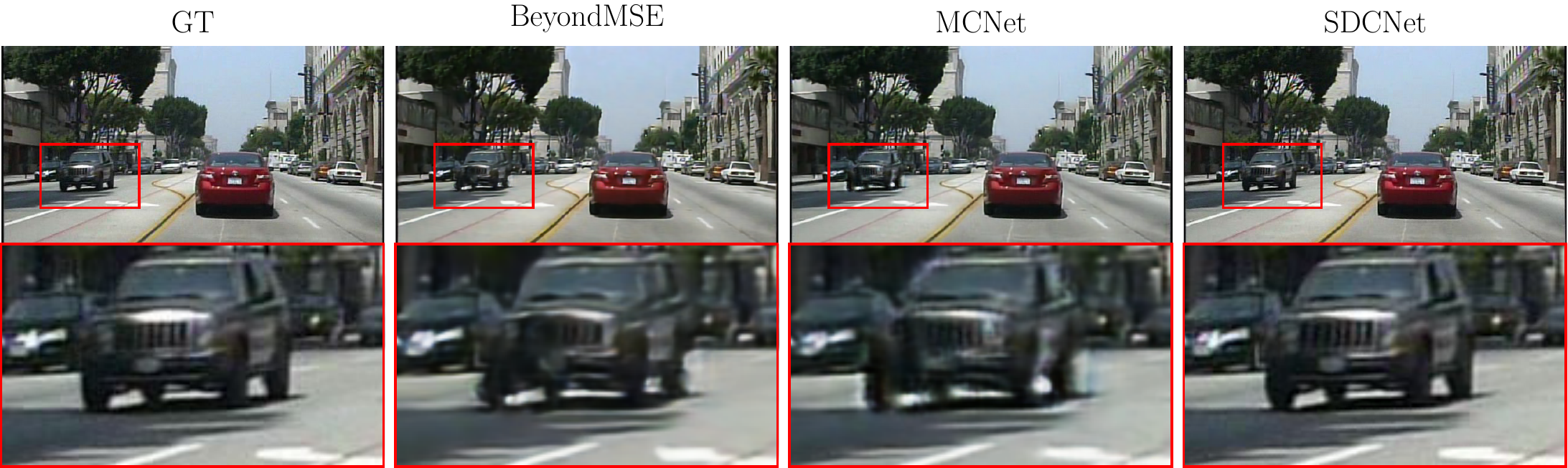}
    \caption{Qualitative comparison for Caltech (set006-v001/506th frame). Left to right: Ground-truth, BeyondMSE, MCNet, and SDC-Net predicted frames.}
    \label{fig:caltech_qual}
\end{figure}

 In Fig. \ref{fig:caltech_qual}, we present qualitative comparisons on CaltechPed. SDC-Net predicted frames are crisp, sharp and show minimal un-natural deformation of the highlighted car (framed in red). All predictions were able in picking up the right motion as shown with their good alignment with the ground-truth frame. However, both BeyondMSE and MCNet create generally blurrier predictions and unnatural deformations on the highlighted car.
\subsection{Comparison on high-definition videos}
Table \ref{table:2} presents next frame prediction comparisons with BeyondMSE, MCNet and CopyLast on 26 full-HD YouTube vidoes. Our SDC-Net model outperforms all other models, achieving an L2 of ${2.4\times{10}^{-3}}$ and SSIM of $0.911$, compared to the state-of-the-art MCNet model which has an L2 of ${2.55\times{10}^{-3}}$ and SSIM of $0.895$.  

\begin{table}[t!]
\caption{Next frame prediction accuracy on YouTube-8M~\cite{abu2016youtube}.}\label{table:2}
\centering
\begin{tabular}{ p{3cm} p{1.5cm} p{1.5cm} p{1.5cm} p{1.5cm}} 
\hline
YouTube8M & L1 & L2 & PSNR & SSIM  \\
\hline 
 BeyondMSE\cite{Mathieu2016BeyondMSE} & 0.0271 &  0.00328 &   33.33 & 0.858 \\
 MCNet\cite{decomposing_motion_content} & 0.0216 & 0.00255 & 35.64 & 0.895  \\
 CopyLast & 0.0260 &   0.00506 &   36.63 & 0.854  \\ 
 Our Vector & 0.0177 & 0.00270    &   37.24 & 0.905  \\ 
 Our Kernel & 0.0186 &   0.00303 &   \textbf{37.33} & 0.904 \\
 Our SDC & \textbf{0.0174} &   \textbf{0.00240} &   37.15 & \textbf{0.911} \\ 
\hline
\end{tabular}
\end{table}

\begin{figure*}[h!]
\centering
\includegraphics[scale=0.7]{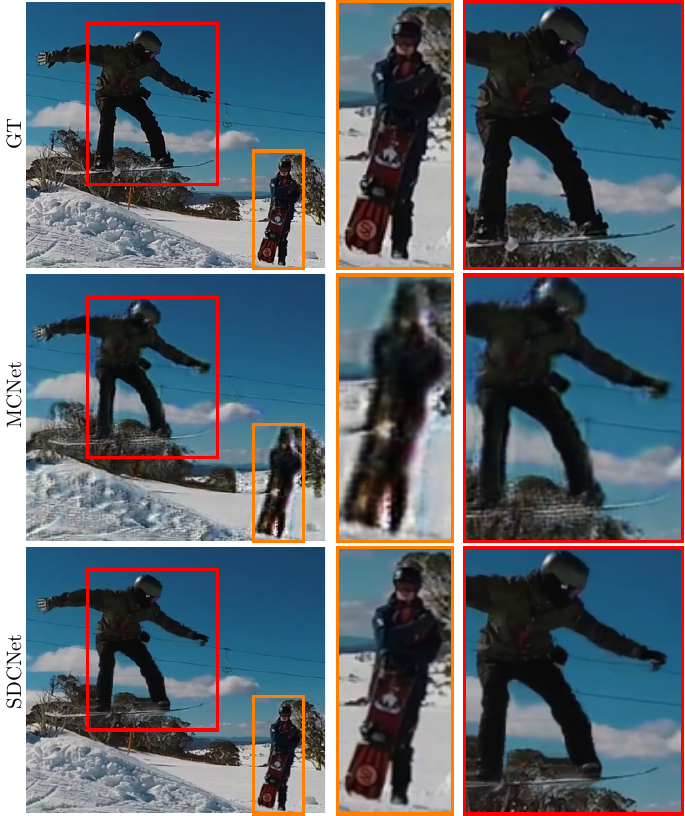}
\caption{Comparison of frame prediction methods. Shown from top to bottom are Ground-truth image, MCNet and SDC-Net results. SDCNet is shown to provide crisp and sharp frames, with motion mostly in good alignment with the ground-truth frame. MCNet results on the other hand appear blurry, with artifacts surrounding the persons (framed in red and orange). MCNet results also show checkerboard artifacts near the skis and on the snow background.}
\label{fig:youtube_motion}
\end{figure*}

In Fig. \ref{fig:youtube_motion}, SDCNet is shown to provide crisp and sharp frames, with motion mostly in good alignment with the ground-truth frame. Since our models do not hallucinate pixels, they produce visually good results by exploiting the image content of the last input frame. For instance, instead of duplicating the borders of foreground objects, our models displace to appropriate locations in the previous frame and synthesize pixels by convolving the learned kernel for that pixel with an image patch centered at the displaced location. 

Since our approach takes FlowNet2~\cite{ilg2017flownet} predicted flows as part of its input, the transformation parameters predicted by our deep model are affected by inaccurate optical flows. For instance, optical flow for the ski in Fig. \ref{fig:youtube_motion} (bottom right) is challenging, and so the ski movement was not predicted by our model as well as the movement of the skiing person.

\begin{figure}
    \centering
    \includegraphics[width=\textwidth]{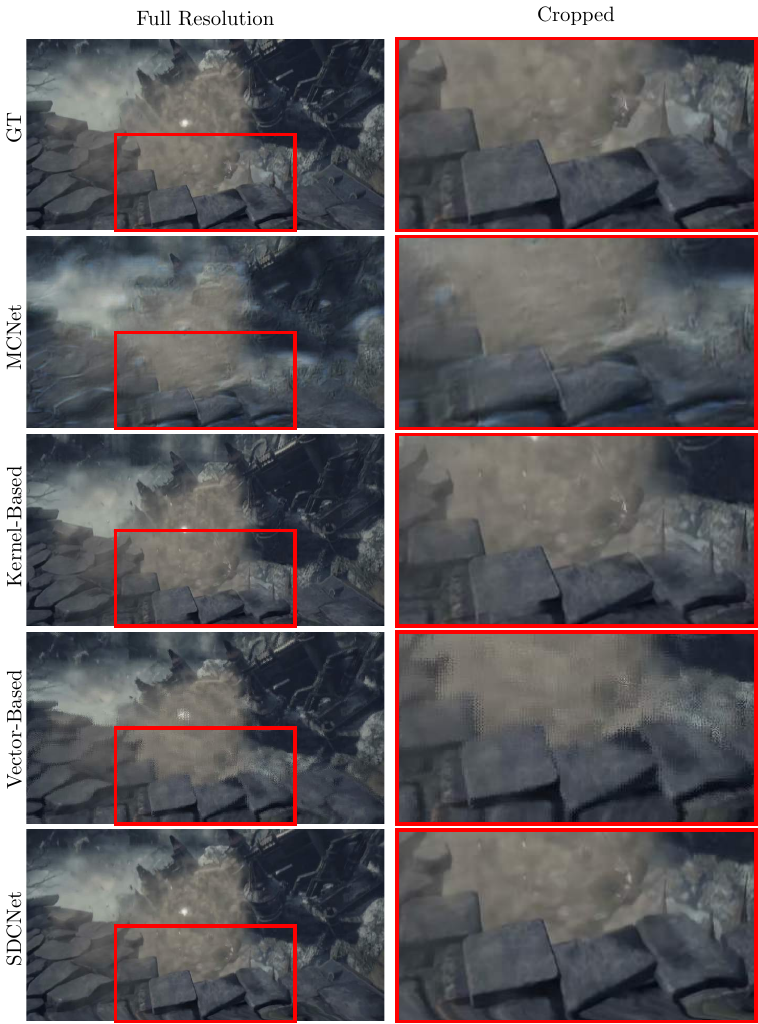}
    \caption{Comparison of frame prediction for large motion. Expected transformation is an upwards displacement with a slight zoom-in. While the Kernel-based, Vector-based, and SDC-based models were all trained with L1 and fine-tuned with style-loss to promote sharpness, note that the Vector-based result still loses coherence when predicting large displacement. On the other hand, the SDCNet is able to displace as much as the Vector-based model while maintaining sharpness. While the Kernel-based result is relatively sharp, it is conservative about predicting the upwards translation (note the relative distance of tiles to the bottom of the frame compared to the vector and SDC approaches). Further, there is a slight ghosting effect in the right-most tile of the Kernel-based result, which is not present in the SDC result.}
    \label{fig:large_motion}
\end{figure}

In Fig. \ref{fig:large_motion}, we qualitatively show comparisons for MCNet, our Kernel-, Vector-, and SDC-based methods for a large camera motion. MCNet shows significantly blurry results and ineffectiveness in capturing large motions. MCNet also significantly alters the color distribution in the predicted frames. Our Kernel-based method has difficulty predicting large motion, but preserves the color distribution. However, the Kernel-based approach often moves components disjointly, leading to visually inferior results. Our Vector-based approach better captures large displacement, such as the motion present in this sequence. However, its predictions suffer from pixel noise. Our SDC-based method, which combines merits of both our Kernel- and Vector-based approaches, combines the ability of our Vector-based method to predict large motions, along with the visually pleasing results of our Kernel-based approach.
\subsection{Comparison in multi-step prediction}
\begin{figure}[!t]
\minipage{0.25\textwidth}
  \includegraphics[width=\linewidth]{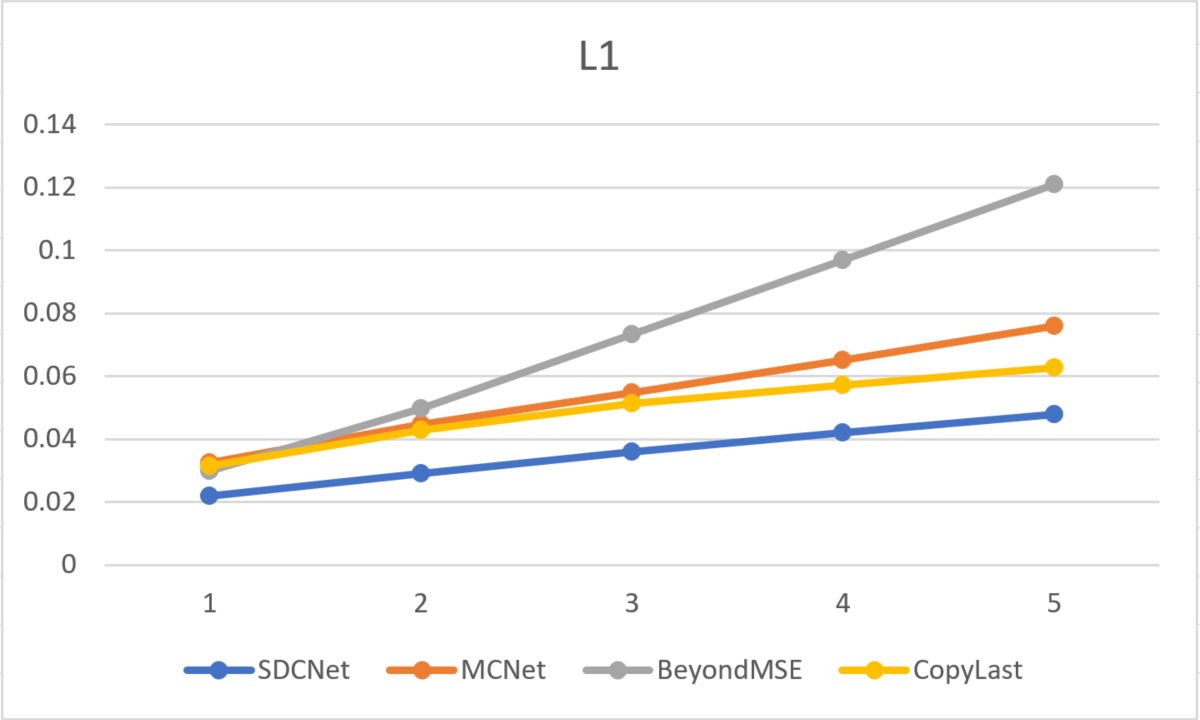}
\endminipage\hfill
\minipage{0.25\textwidth}
  \includegraphics[width=\linewidth]{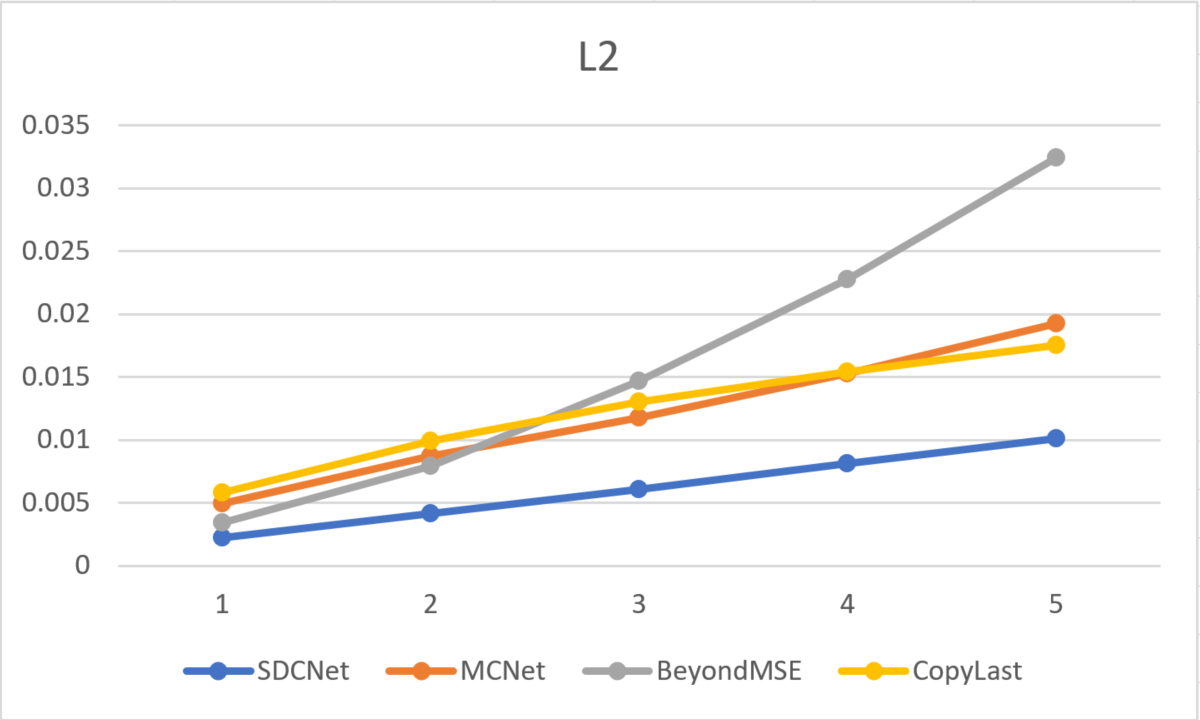}
\endminipage\hfill
\minipage{0.25\textwidth}%
  \includegraphics[width=\linewidth]{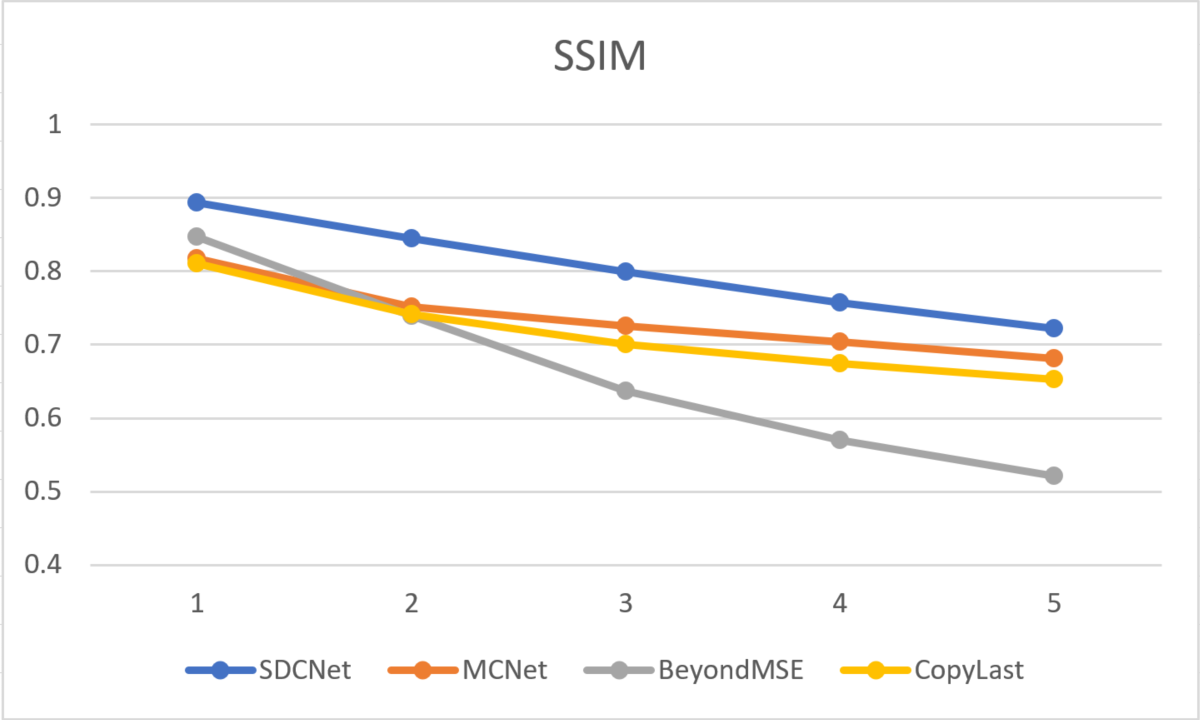}
\endminipage
\minipage{0.25\textwidth}%
  \includegraphics[width=\linewidth]{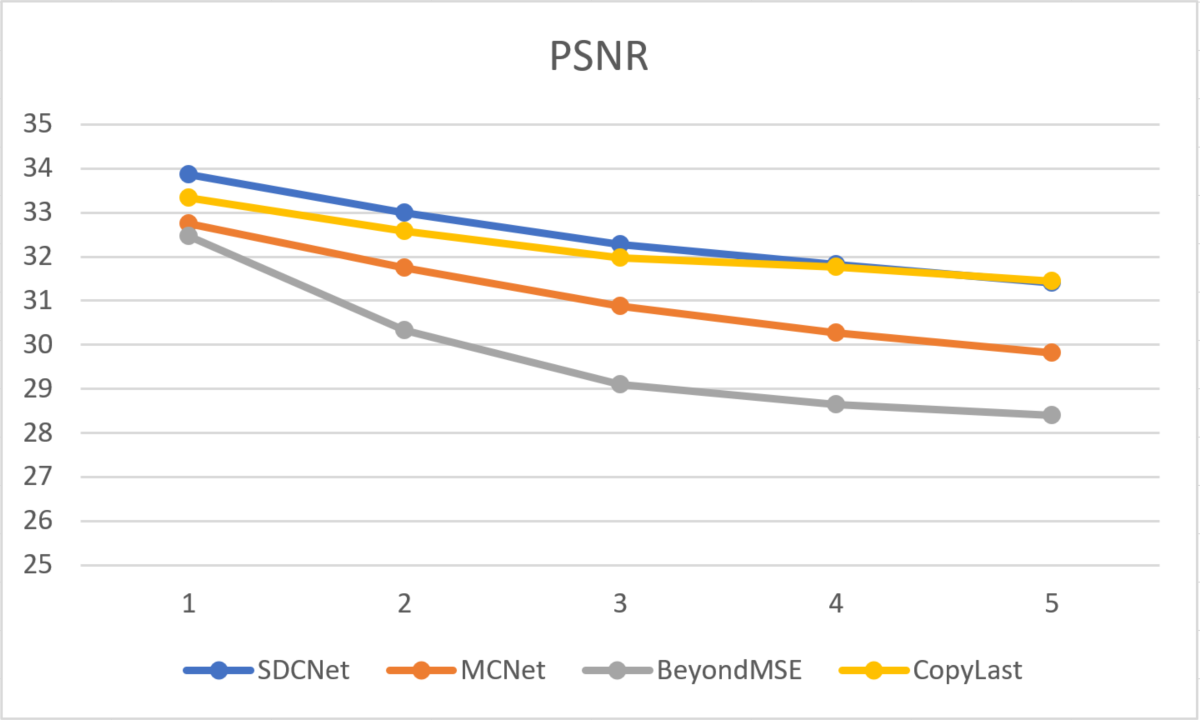}
\endminipage
\caption{Quantitative \textit{five-step} prediction results for SDC-Net (blue), MCNet (orange), BeyondMSE (gray) and CopyLast (yellow). SDCNet shows consistently better results as quantified by L1, L2, PSNR and SSIM over 120,725 unique CaltechPed frames.}
\label{fig:quant}
\end{figure}

\begin{figure}[!htb]
\minipage{0.2\textwidth}
  \includegraphics[width=\linewidth]{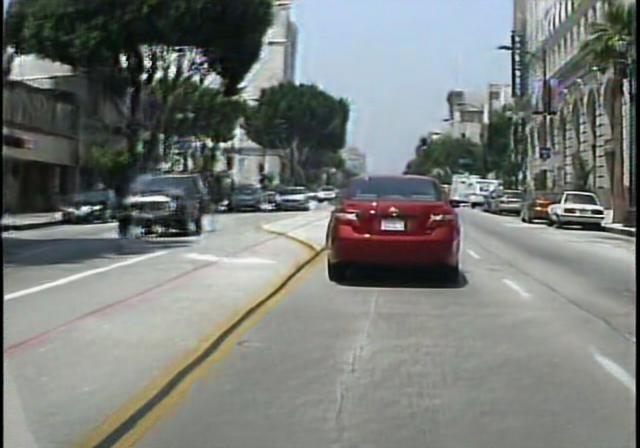}
\endminipage\hfill
\minipage{0.2\textwidth}
  \includegraphics[width=\linewidth]{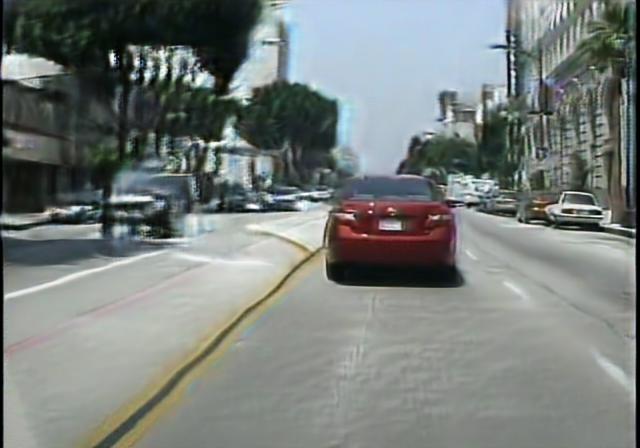}
\endminipage\hfill
\minipage{0.2\textwidth}%
  \includegraphics[width=\linewidth]{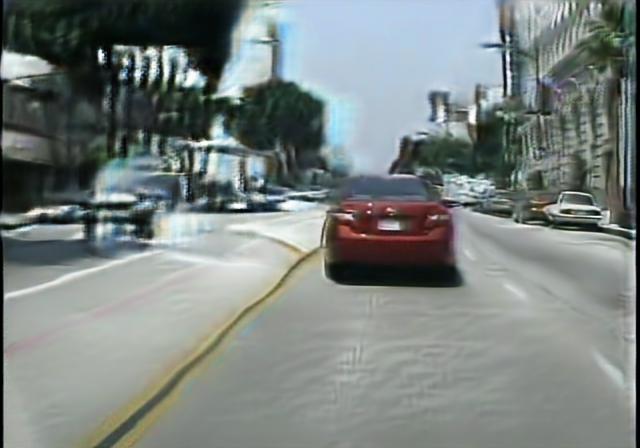}
\endminipage
\minipage{0.2\textwidth}%
  \includegraphics[width=\linewidth]{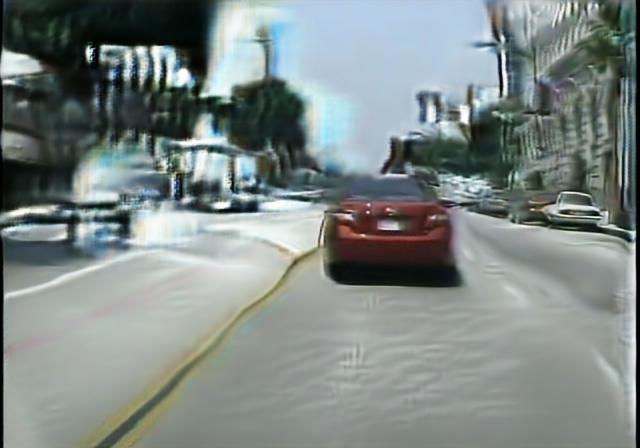}
\endminipage
\minipage{0.2\textwidth}%
  \includegraphics[width=\linewidth]{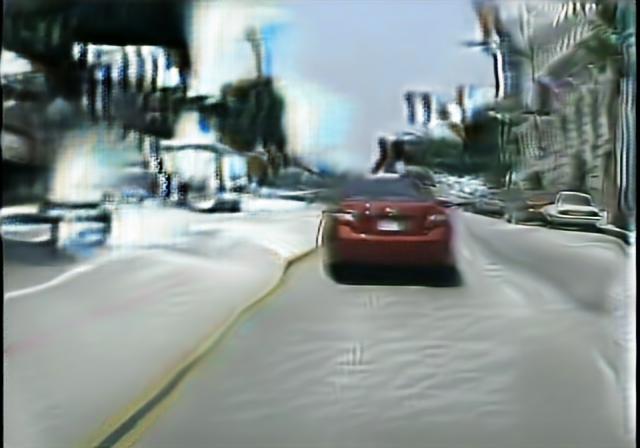}
\endminipage
\newline
\minipage{0.2\textwidth}
  \includegraphics[width=\linewidth]{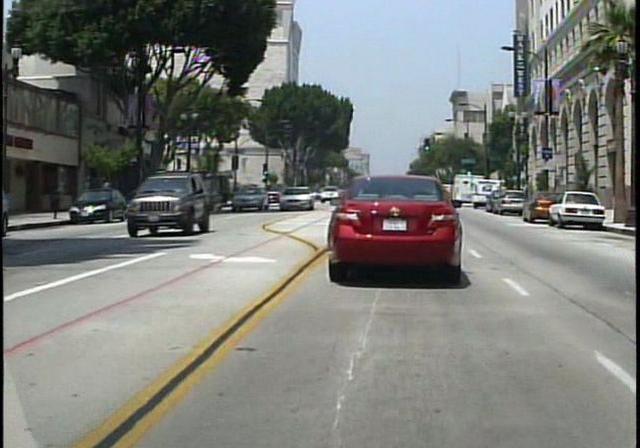}
\endminipage\hfill
\minipage{0.2\textwidth}
  \includegraphics[width=\linewidth]{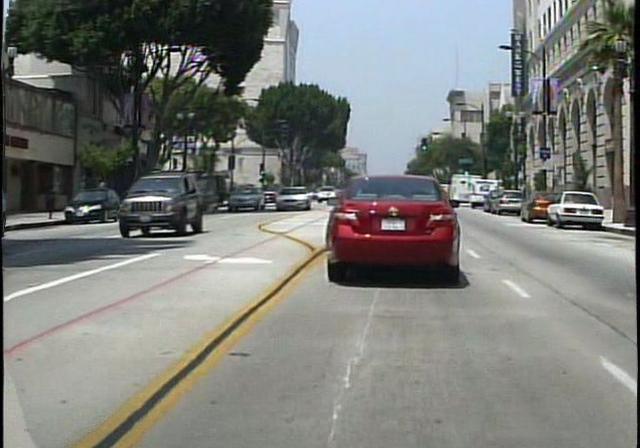}
\endminipage\hfill
\minipage{0.2\textwidth}%
  \includegraphics[width=\linewidth]{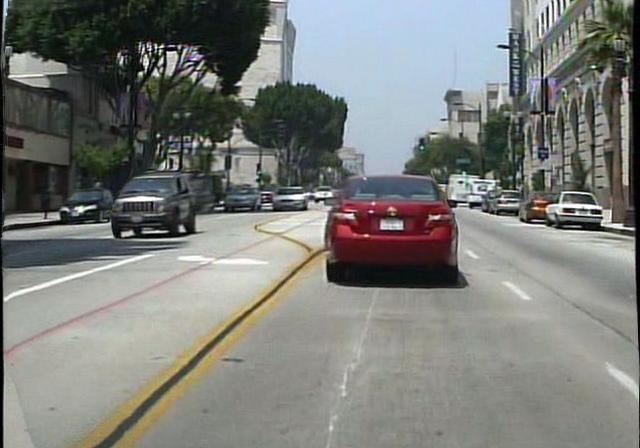}
\endminipage
\minipage{0.2\textwidth}%
  \includegraphics[width=\linewidth]{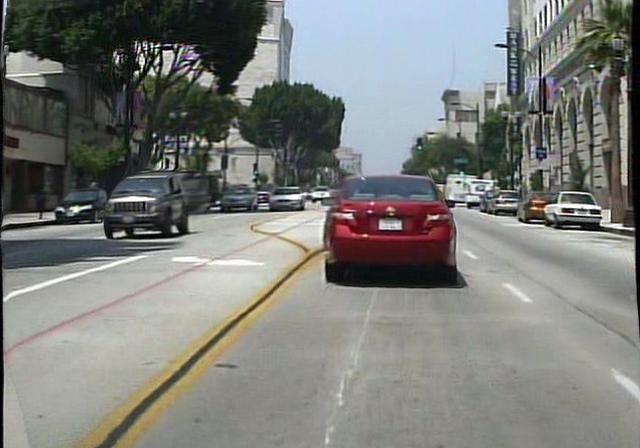}
\endminipage
\minipage{0.2\textwidth}%
  \includegraphics[width=\linewidth]{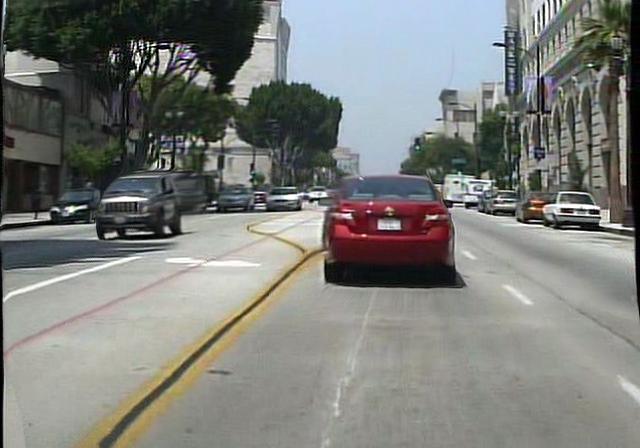}
\endminipage
\newline
\minipage{0.2\textwidth}
  \includegraphics[width=\linewidth]{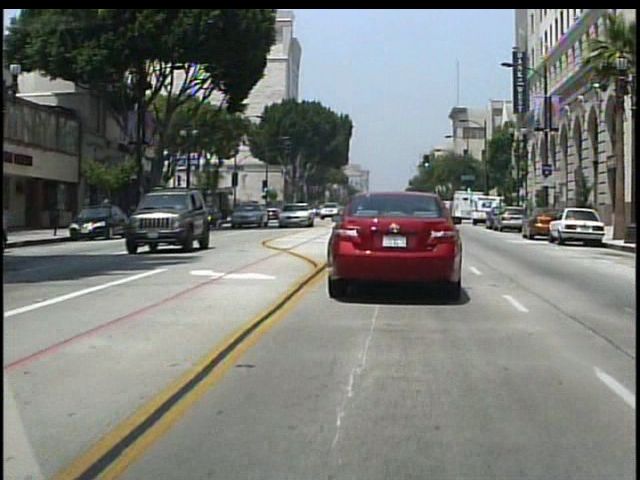}
\endminipage\hfill
\minipage{0.2\textwidth}
  \includegraphics[width=\linewidth]{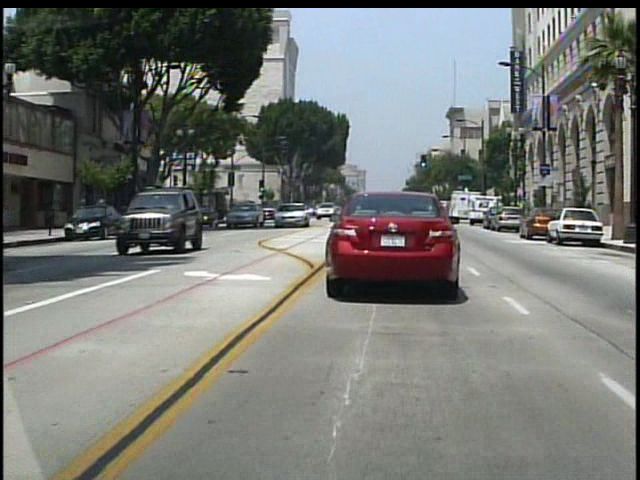}
\endminipage\hfill
\minipage{0.2\textwidth}%
  \includegraphics[width=\linewidth]{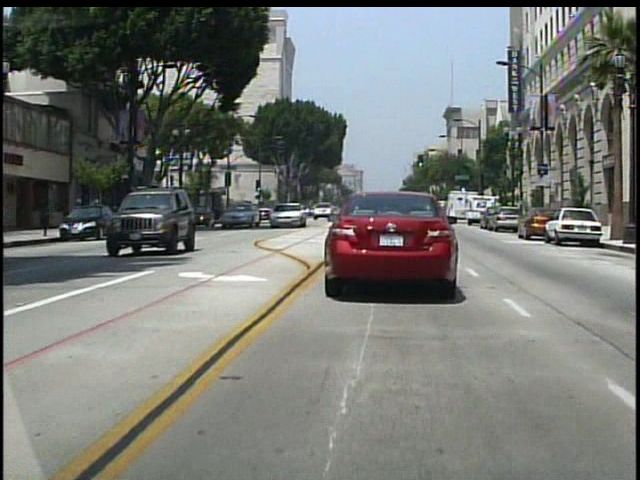}
\endminipage
\minipage{0.2\textwidth}%
  \includegraphics[width=\linewidth]{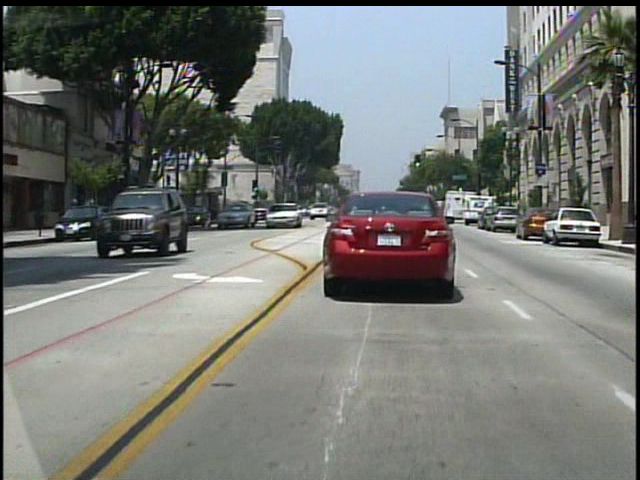}
\endminipage
\minipage{0.2\textwidth}%
  \includegraphics[width=\linewidth]{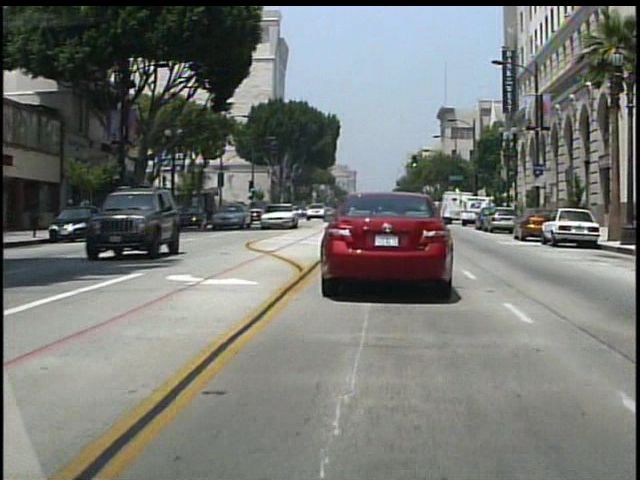}
\endminipage
\caption{Qualitative \textit{five-step} prediction results for MCNet (top row), SDCNet (middle row), and Ground Truth (bottom row). Both MCNet and SDCNet were conditioned on the same set of five frames (not seen in the figure).}
\label{fig:qual}
\end{figure}

Previous experiments showed SDCNet's performance in next frame prediction. In practice, models are used to predict multiple future frames. Here, we condition each approach on five original frames and predict five future frames on CaltechPed. Fig. \ref{fig:quant} shows that SDCNet predicted multiple frames are consistently favourable when compared to previous approaches, as quantified by L1, L2, SSIM and PSNR over 120,725 unique Caltech Pedestrian frames. Fig. \ref{fig:qual} presents an example five-step prediction that show SDCNet predicted frames preserving color distribution, object shapes and their fine details. 

\subsection{Ablation results}

\begin{figure*}[h!]
\centering
\includegraphics[width=\textwidth]{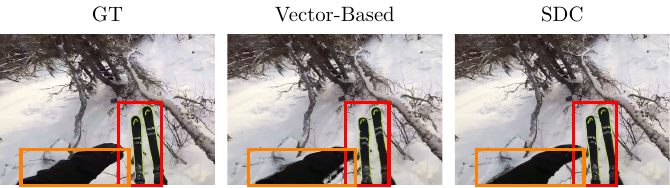}
\caption{Comparison of frame synthesis operations. Ground-truth frame (left), Vector-based sampling (middle), and SDC (right). Some foreground duplication (orange box) and inconsistent pixel synthesis (red box, may require zooming in) are present in the Vector-based approach but resolved in the SDC results.}
\label{fig:flow_vs_transfomer}
\end{figure*}

We compare our Vector-based with our SDC-based approach in Fig. \ref{fig:flow_vs_transfomer}. Our Vector-based approach struggles with disocclusions (orange box), as described in \ref{sec:optical_flow}. In Fig. \ref{fig:flow_vs_transfomer}, the Vector-based model avoids completely stretching the glove borders, but still leaves some residual glove pixels behind. The Vector-based approach also may produce speckled noise patterns due to large motion (red box). Disocclusion and speckled noise are significantly reduced in the SDC-Net results shown in Fig.\ref{fig:flow_vs_transfomer}. 

\begin{figure*}[t!]
\centering
\includegraphics[scale=0.38]{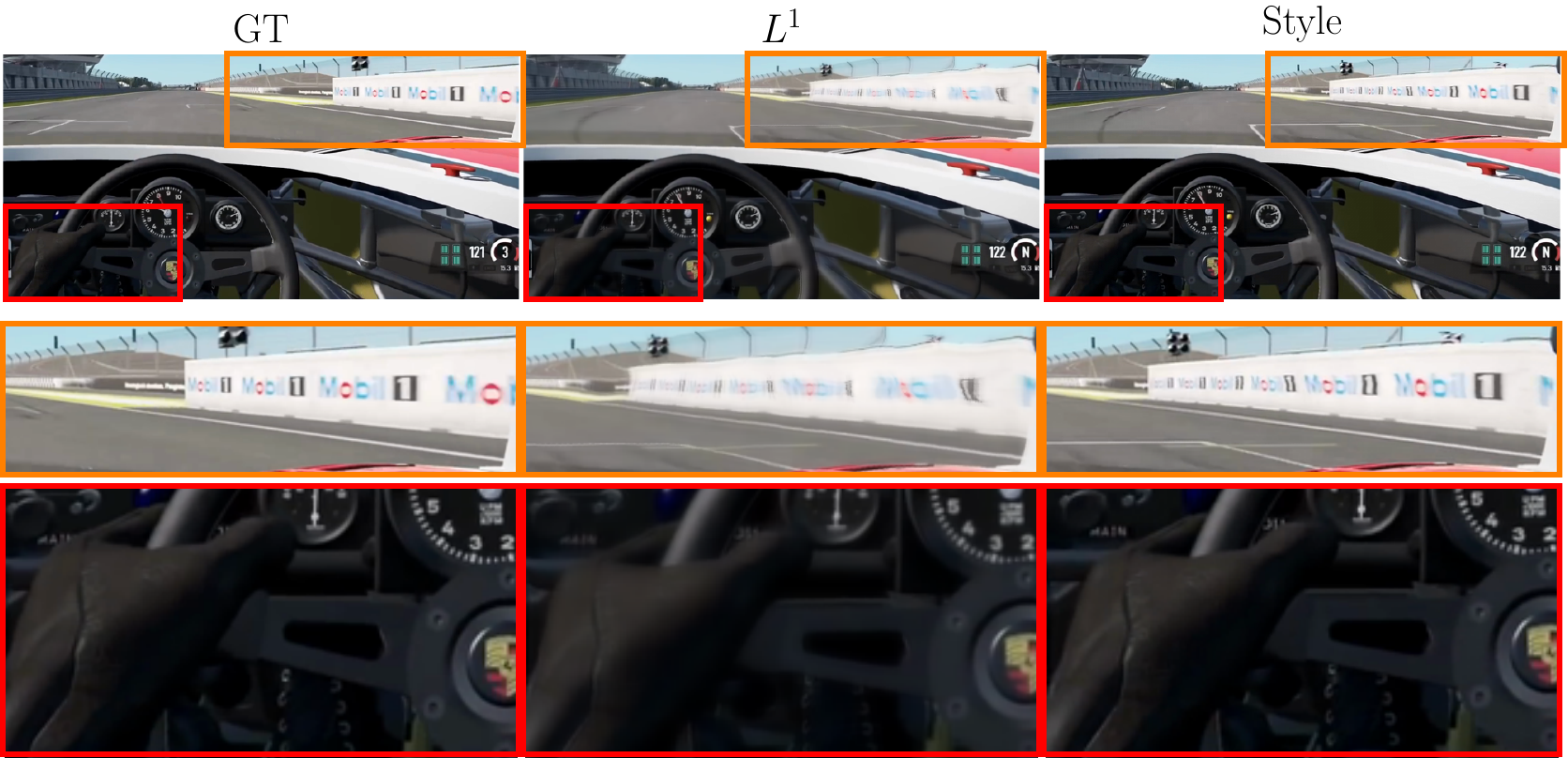}
\caption{Comparison of loss functions. Ground-truth (left), L1 loss (middle), and Fine-tuned result with style loss (right). Fine-tuning with style loss can improve the sharpness of results, as seen in the rendered text on the barriers and fence (orange crop) as well as the glove (red crop).}
\label{fig:l1_vs_style}
\end{figure*}

In Fig. \ref{fig:l1_vs_style}, we present qualitative results for our SDC-based model trained using ${\mathcal{L}}_{1}$ loss alone vs ${\mathcal{L}}_{1}$ followed by our ${\mathcal{L}}_{finetune}$ given by equation (\ref{eq:finetune}). We note that using ${\mathcal{L}}_{1}$ loss alone leads to slightly blurry results, e.g. the glove (red box), and the fence (orange box) in Fig.\ref{fig:l1_vs_style}. Fig. \ref{fig:l1_vs_style} (center column)
shows the same result after fine-tuning, with finer details preserved -- demonstrating that the perceptual and style losses reduce blurriness. We also observed that the ${\mathcal{L}}_{1}$ loss helps capture large
motions that are otherwise challenging to capture.

Fig. \ref{fig:l1_vs_style} represents a challenging example due to fast motion. Since our model depends on optical flow, situations that are challenging for optical flow are also difficult for our model. The prediction errors can be seen with the relatively larger misalignment on the fence compared to the ground truth (orange box). Our approach also fails during scene transitions, where past frames are not relevant to future frames. Currently, we automatically detect scene transitions by analyzing optical flow statistics, and skip frame prediction until enough (five) valid frames to condition our models are available.

\section{Conclusions}
We present a 3D CNN and a novel spatially-displaced convolution (SDC) module that achieves state-of-the-art video frame prediction. Our SDC module effectively handles large motion and allows our model to predict crisp future frames with motion closely matching that of ground-truth sequences. We trained our model on 428K high-resolution video frames collected from gameplay footage. To the best of our knowledge, this is the first attempt in transfer learning from synthetic to real life for video frame prediction. Our model's accuracy is dependent on the accuracy of the input estimated flows, thus leading to failures in fast motion sequences. Future work will include a study on the effect of multi-scale architectures for fast motion.

\textbf{Acknowledgements}: We would like to thank Jonah Alben, Paulius Micikevicius, Nikolai Yakovenko, Ming-Yu Liu, Xiaodong Yang, Atila Orhon, Haque Ishfaq and NVIDIA Applied Research staff for suggestions and discussions, and Robert Pottorff for capturing the game datasets used for training.

\clearpage
%
%
%
\bibliographystyle{splncs04}
\bibliography{bibliography}
\end{document}